\documentclass{article}
\usepackage{ijcai22}

\usepackage{times}
\usepackage{soul}
\usepackage{url}
\usepackage{hyperref}
\usepackage[utf8]{inputenc}
\usepackage[small]{caption}
\usepackage{graphicx}
\usepackage{amsmath}
\usepackage{amsthm}
\usepackage{booktabs}
\usepackage{algorithm}
\usepackage{algorithmic}
\usepackage{multirow} % add by jzk
\usepackage{color,xcolor} % add by jzk
\usepackage{pifont} % add by jzk
\usepackage{booktabs} % add by jzk
\usepackage{amsmath} % add by jzk
\usepackage{amssymb} % add by jzk
\usepackage{bbm} % add by jzk
\usepackage{bm} % add by jzk
\usepackage{textcomp} % add by jzk
\usepackage{algorithm} % add by jzk
\usepackage{algorithmic} % add by jzk
\usepackage{colortbl} % add by jzk
\usepackage{bbding} % add by jzk
\usepackage{booktabs} % add by jzk 
\usepackage{pifont} % add by jzk

\usepackage{algorithm}
\usepackage{algorithmic}
\usepackage{newfloat}
\usepackage{listings}
\floatstyle{ruled}
\newfloat{listing}{tb}{lst}{}
\floatname{listing}{Listing}
\title{DiffusionInst: Diffusion Model for Instance Segmentation}

\author{
Zhangxuan Gu$^1$
\and
Haoxing Chen$^{1,2}$\and
Zhuoer Xu$^1$\and
Jun Lan$^1$ \\
Changhua Meng$^1$\and
Weiqiang Wang$^1$
\affiliations
$^1$Tiansuan Lab, Ant Group Inc.\\
$^2$Nanjing University
\emails
 \{guzhangxuan.gzx,xuzhuoer.xze,yelan.lj,changhua.mch,weiqiang.wwq\}@antgroup.com haoxingchen@smail.nju.edu.cn}

\begin{document}

\maketitle

\begin{abstract}
Diffusion frameworks have achieved comparable performance with previous state-of-the-art image generation models. Researchers are curious about its variants in discriminative tasks because of its powerful noise-to-image denoising pipeline. This paper proposes DiffusionInst, a novel framework that represents instances as instance-aware filters and formulates instance segmentation as a noise-to-filter denoising process. The model is trained to reverse the noisy groundtruth without any inductive bias from RPN.
During inference, it takes a randomly generated filter as input and outputs mask in one-step or multi-step denoising. Extensive experimental results on COCO and LVIS show that DiffusionInst achieves competitive performance compared to existing instance segmentation models with various backbones, such as ResNet and Swin Transformers. We hope our work could serve as a strong baseline, which could inspire designing more efficient diffusion frameworks for challenging discriminative tasks. Our code is available in \textit{\color{magenta}{\url{https://github.com/chenhaoxing/DiffusionInst}}}.
\end{abstract}

\section{Introduction}

Instance segmentation aims to represent objects with binary masks, which is a finer-grained representation compared to the bounding boxes of object detection. Standard instance segmentation approaches can be divided into two groups, {\em i.e.,} two-stage\cite{maskrcnn,PANet,maskscoringrcnn}, and single-stage\cite{BlendMask,YOLACT,SOLO,SOLOv2,CondInst}. Two-stage methods first detect objects, then crop their region features with RoI alignment to further classify each pixel. At the same time, the framework of single-stage instance segmentation is usually based on anchors and is thus much simpler. However, they all have dense prediction heads, requiring the non-maximum suppression (NMS) technique during inference.

Recently, SOLQ\cite{dong2021solq}, QueryInst\cite{QueryInst} and Mask2Former\cite{Mask2Former} proposed end-to-end instance segmentation frameworks with the help of learnable queries and bipartite matching. Specifically, they extend DETR\cite{DETR} by feeding the instance-aware RoI features to the mask head for predicting instance masks. Unlike existing anchor-based and anchor-free methods, query-based approaches use randomly generated queries to replace the RPN and anchors, reducing the inductive bias in localization instances and improving the segmentation performance by one-to-one label assignment.

Considering that query-based approaches\cite{Mask2Former,QueryInst} formulate like a noise-to-mask scheme, we believe they are a special case of diffusion models\cite{DDPM,SGM,GMDD}. To be exact, they directly denoise random queries to objects by only one forward pass of their decoders, while diffusion models can additionally perform multi-step denoising gradually during inference. It inspired us to explore a new framework for instance segmentation with the diffusion process.

\begin{figure}[t]
\centering
\includegraphics[width=0.49\textwidth]{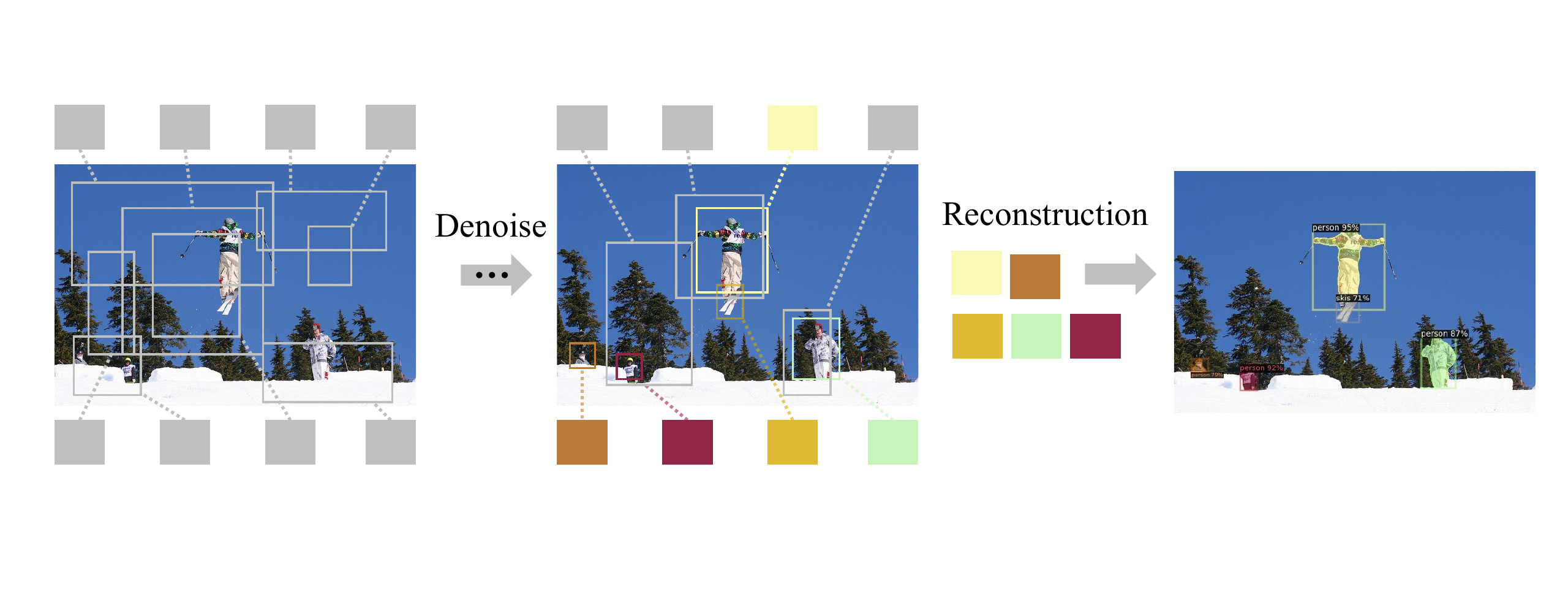}
\caption{\textbf{Diffusion model for instance segmentation.} We propose to regard instance segmentation as a denoising diffusion process from noisy 
bounding boxes and filters to instance masks with a dynamic mask head for mask reconstruction.}
\label{intro}
\end{figure}

However, how to adapt the diffusion model in instance segmentation is still an open problem. Recently, DiffusionDet\cite{DiffusionDet} has been proposed to tackle the object detection task by casting detection as a generative task over the space of bounding boxes in the image. At the training stage, it adds Gaussian noise to groundtruth bounding boxes to obtain noisy boxes. Then the RoI features of noisy boxes are fed to the decoder for predicting groundtruth boxes. The whole network works like a denoising pipeline. During inference, DiffusionDet generates bounding boxes by iteratively feeding random initialized boxes to the decoder network as the reverse of the diffusion process. 

According to CondInst\cite{CondInst}, instance masks within one image can be represented by instance-aware filters (vectors) with a common mask feature. Inspired by it, this paper proposes DiffusionInst, a novel instance segmentation framework from a noise-to-filter diffusion view. By reusing the pipeline of DiffusionDet, we have made two changes to the instance segmentation task. Firstly, besides bounding boxes, we also generate noisy filters during diffusion. Secondly, we introduce a mask branch to obtain multi-scale information from FPN\cite{FPN} for global mask reconstruction. We show the denoising diffusion process of DiffusionInst in Figure~\ref{intro}.

Besides the ability to perform multi-step inference, another advantage of DiffusionInst compared to query-based models during training is that our noisy generated filters may contain different distribution noises conditioned on the randomly chosen time $t\in \{0,1,\cdots,T\}$. In some cases, $T$ denoising times/steps can be viewed as $T$ different distribution noises, which significantly increases the difficulty of model learning and contributes much to model robustness and performance.

To this end, we evaluate DiffusionInst on COCO\cite{COCO} validation dataset. With ResNet-$50$\cite{ResNet} backbone, DiffusionInst achieves 37.3\% AP using one-step denoise, which significantly outperforms Mask RCNN\cite{maskrcnn} (34.4\% AP), SOLO\cite{SOLO} (33.9\% AP), and CondInst\cite{CondInst}(35.9\% AP). Besides, we can further improve DiffusionInst up to 47.8\% AP by employing larger model Swin Transformers\cite{swin} as the backbone, which achieves better performance than QueryInst\cite{QueryInst}(44.6\% AP). Similar conclusions can be drawn on COCO test-dev and long-tailed LVIS\cite{LVIS} dataset.
%The differences between standard, query-based and diffusion-based instance segmentation approaches are shown in Figure~\ref{}.

In summary, our main contributions are: 
\begin{itemize}
%%%%%%%%%%%%%%%%%%%%%%%%%%
    \item We propose DiffusionInst, the first work of diffusion model for the instance segmentation by regarding it as a generative noise-to-filter diffusion process. 
    \item Instead of predicting local masks, we utilize instance-aware filters and a common mask branch feature to represent and reconstruct global instance masks.
    \item Comprehensive experiments are conducted on the COCO and LVIS benchmarks. DiffusionInst achieves competitive results compared with existing well-designed approaches, showing the promising future of diffusion models in discriminative tasks.
%%%%%%%%%%%%%%%%%%%%%%%%%%%%
\end{itemize}

\section{Related Works}
\subsection{Instance Segmentation}
Instance segmentation aims to predict pixel-wise instance masks with class labels for each instance presented in each image. The existing methods can be roughly summarized into some categories. Top-down methods\cite{LiFCIASS,maskrcnn,PANet,BlendMask} detect the object first and then segment the object in the box. Bottom-up methods\cite{SGN,SSAP,AE} learn the pixel-wise embeddings and then cluster them into groups. Direct methods\cite{SOLO,SOLOv2} perform instance segmentation directly without box detection or embedding learning. More recently, SOLQ\cite{dong2021solq}, QueryInst\cite{QueryInst} and Mask2Former\cite{Mask2Former} proposed to decode random queries to objects for end-to-end instance segmentation frameworks with the success of DETR\cite{DETR} in the object detection task. Unlike the above methods, we are the first to formulate instance segmentation as a generative denoising process with competitive performances.

\subsection{Diffusion Model}
Diffusion model\cite{DDPM,SGM,GMDD} is a parameterized Markov chain, which starts from the sample in random distribution and reconstructs the data sample via a gradual denoising process. Recently, diffusion models have made remarkable achievements in many fields, e.g., computer vision\cite{CascadedDiff,LDM,LDEM,3DDiff}, language understanding\cite{DiffLM,D3PM,Diffuseq}, robust learning\cite{GDM_Adv,DiffPure} and temporal data modeling\cite{NMC_SDE,DiffWave}.

\subsection{Diffusion Model for Visual Understanding.}
Diffusion models have achieved great success in image generation and synthesis\cite{Prafullanips,DDPM,GMDD}. However, their potential for visual understanding has yet to be fully explored. Recently, Chen \textit{et al.}\cite{DiffSeg} adopted analog bits based diffusion model\cite{BitDiff} to model panoptic masks. Chen \textit{et al.}\cite{DiffusionDet} formulated object detection as a noise-to-box task. In this paper, we further broaden the application of the diffusion model by formalizing instance segmentation as a denoising process. To the best of our knowledge, this is the first work that adopts a diffusion model for the instance segmentation task.

\section{Methodology}
In this section, we first briefly review the pipeline of diffusion models and DiffusionDet\cite{DiffusionDet}. Then, we introduce different instance mask representation methods. Next, we present the architecture of DiffusionInst and its training and inference process. At last, we provide some discussions about employing the diffusion model in instance segmentation.
\begin{figure*}[t]
\centering
\includegraphics[width=0.95\textwidth]{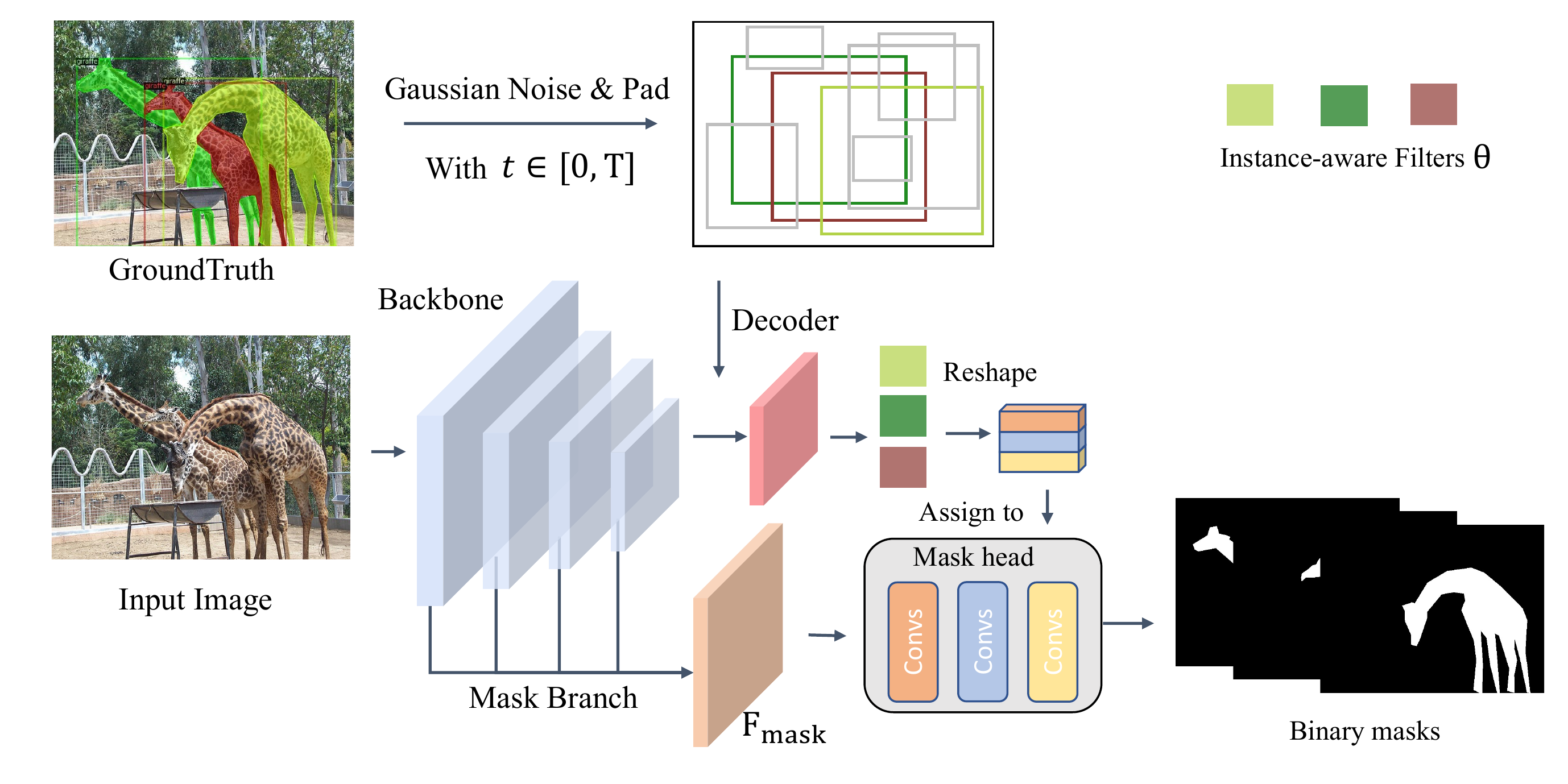}
\caption{\textbf{The overview of our DiffusionInst.} The backbone with FPN extracts multi-scale features from an input image. During training, we add random $t$ step noise to the groundtruth boxes and pad them to predefined numbers. Instance-aware noisy filters are constructed by combining features and noisy boxes. We additionally develop a mask branch to keep multi-scale information in $\mathbf{F}_{mask}$. By applying convolutions whose weights are assigned from noisy filters $\bm{\theta}$ to $\mathbf{F}_{mask}$, we can reconstruct instance masks. In inference, the noisy filters are randomly sampled from the Gaussian distribution. Note that input images will only go through the backbone and mask branch once while the multi-step denoising process is performed on the boxes and filters.}
\label{framework}
\end{figure*}

\subsection{Preliminaries}

\noindent \textbf{Diffusion Model:} Recent diffusion models usually use two Markov chains: a forward chain that perturbs the image to noise and a reverse chain that refines noise back to the image. Formally, given a data distribution $\mathbf{x}_0 \sim q(\mathbf{x}_0)$, the forward noise perturbing process at time $t$ is defined as $q(\mathbf{x}_t|\mathbf{x}_{t-1})$. It gradually adds Gaussian noise to the data according to a variance schedule $\beta_1,\cdots,\beta_T$:
\begin{equation}\label{eqn.forward}
q(\mathbf{x}_t|\mathbf{x}_{t-1}) = \mathcal{N}(\mathbf{x}_t;\sqrt{1-\beta_t}\mathbf{x}_{t-1},\beta_t\mathbf{I}).
\end{equation}
Given $\mathbf{x}_0$, we can easily obtain a sample of $\mathbf{x}_t$ by sampling a Gaussian vector $\mathbf{\epsilon} \sim \mathcal{N}(\mathbf{0},\mathbf{I})$ and applying the transformation as follows:
\begin{equation}\label{eqn.forward}
\mathbf{x}_t = \sqrt{\bar{\alpha_t}}\mathbf{x}_0+(1-\bar{\alpha_t})\mathbf{\epsilon},
\end{equation}
where $\bar{\alpha_t} = \prod^t_{s=0}(1-\beta_s)$.

During training, a neural network is trained to predict $\mathbf{x}_0$ from $\mathbf{x}_t$ for different $t\in \{1,\cdots,T\}$. While performing inference, we start from a random noise $\mathbf{x}_T$ and iteratively apply the reverse chain to obtain $\mathbf{x}_0$. We refer the readers to \cite{yang2022diffusion} for more details.

\noindent \textbf{DiffusionDet:} It is the first diffusion model in the object detection task. In their setting, data samples are a set of bounding boxes $\mathbf{x}_0=\mathbf{b}$, where $\mathbf{b}\in \mathcal{R}^{N\times 4}$ is a set of $N$ boxes. 

During training, DiffusionDet first constructs the diffusion process and then reverses this process. By padding extra boxes to the original groundtruth boxes, the model can handle a fixed number of instance boxes. Set prediction loss\cite{DETR} is utilized to optimize the whole DiffusionDet with optimal transport assignment\cite{DBLP:journals/corr/abs-2103-14259} as the label assignment strategy.

The inference procedure of DiffusionDet additionally uses DDIM\cite{DBLP:journals/corr/abs-2010-02502} to refine the boxes for the next step in the iterative sampling process. 

\subsection{Mask Representation}

Intuitively, instance masks are usually represented by binary figures. However, according to PolarMask\cite{PolarMask} and BlendMask\cite{BlendMask}, there are various representation methods for an instance mask. For example, PolarMask formulates an instance mask with polar coordinates. It represents one mask with a 36-dim vector from the center point by dividing $360^{\circ}$ into 36 directions, and each value indicates the half-line length. In some cases, boxes (4-dim vectors) can even be viewed as very coarse masks. 

As a result, we use the dynamic mask head to represent instance masks following CondInst\cite{CondInst}. Specifically, instance mask can be generated by convolving an instance-agnostic mask feature map $\mathbf{{F}}_{mask}$ from mask branch and instance-specific filter $\bm{\theta} \in  \mathcal{R}^{d}$, which is calculated as follows:
\begin{equation}\label{maskhead}
    \mathbf{m} = \mathcal{\phi}(\mathbf{{F}}_{mask};\bm{\theta}),
\end{equation}
where $\mathbf{{F}}_{mask}$ is multi-scale fused feature map from FPN features $\{P_3, P_4, P_5\}$. $\mathbf{m} \in \mathbb{R}^{H\times W}$ is the predicted binary mask. The $\mathcal{\phi}$ indicates the mask head, which consists of three $1\times1$ convolutional layers with filter $\bm{\theta}$ as convolution kernel weights. For example, if the mask head $\mathcal{\phi}$ have three convs with channel $\{8,8,1\}$, then the dimension of filters $\bm{\theta}$ is $d = (8\times 8+8\times 8+8) +(8+8+1) = 153$.

\begin{table*}[thb]
\centering
%\resizebox{0.98\textwidth}{!}{
 \begin{tabular}{c|c|c|c|cc|ccc|c}
\toprule 
  \multirow{2}{*}{Methods}
  & \multirow{2}{*}{Backbone}
  & \multirow{2}{*}{Sched.}
  & \multicolumn{6}{c|}{COCO}
  & \multirow{2}{*}{FPS} \\
  %& \multicolumn{6}{c}{LVIS} \\ 
  & &
  & AP 
  & AP$_{\mathtt{50}}$ 
  & AP$_{\mathtt{75}}$ 
  & AP$_{\mathtt{S}}$ 
  & AP$_{\mathtt{M}}$
  & AP$_{\mathtt{L}}$
  &\\
  %& AP 
  %& AP$_{\mathtt{50}}$ 
  %& AP$_{\mathtt{75}}$ 
  %& AP$_{\mathtt{S}}$ 
  %& AP$_{\mathtt{M}}$
  %& AP$_{\mathtt{L}}$ \\
  \midrule
  Mask RCNN &  ResNet-$50$ &1x &34.4 &55.1&   36.7 &   18.1&  37.5 &47.4& 14.0\\
  Cascade Mask RCNN &  ResNet-$50$ &1x & 35.9  &56.6 & 38.4  & 19.4 & 38.5 &49.3&10.4 \\
  SOLO &  ResNet-$50$ &1x &33.9  & 54.1   &35.9  &12.6   & 37.1   &51.4& 13.0 \\
  %SOLOv2 &  ResNet-$50$ &3x & 388  &599   & 417 & 165 & 417  &  562&  &  &  &  & \\
  CondInst &  ResNet-$50$ &1x &35.9  & 56.9&  38.3  &  19.1  & 38.6  & 46.8& 14.1\\
  Mask RCNN &  ResNet-$50$ &3x &37.5& 59.3& 40.2& 21.1& 39.6 &48.3 & 14.0\\
  %Cascade Mask RCNN &  ResNet-$50$ &3x & 38.6 &60.0 &41.7 &21.7 &40.8& 49.6&10.4 \\
  Mask2Former & ResNet-$50$ &4x &43.7  & - & - & 23.4 &47.2 &64.8  &9.7\\
  %Mask DINO& ResNet-$50$ &4x &  &  &  &  &  &  &  &  &  &  &\\
  DiffusionInst &  ResNet-$50$ &1x &30.4 &55.2 & 29.9  & 14.4  & 32.7 &45.3& 1.7\\
  DiffusionInst &  ResNet-$50$ &3x &35.6&58.2&37.4&17.2&38.4&53.1&1.7 \\
  DiffusionInst &  ResNet-$50$ &5x & 37.3&60.3&39.3&18.9&40.1&54.7& 1.7\\
  DiffusionInst(4-step) &  ResNet-$50$ &5x & 37.5&60.9&39.3&19.2&40.4&54.8&1.7  \\\midrule
  Mask RCNN &  ResNet-$101$ &3x & 38.5  & 60.0  & 41.6  & 19.2  & 41.6  & 55.8& 10.8\\
  Cascade Mask RCNN &  ResNet-$101$ &3x & 39.6 &  61.0  &42.8  &19.6    &42.7   &56.8&8.7  \\
  SOLO &  ResNet-$101$ &3x &37.8 &59.5  & 40.4  & 16.4   &40.6    & 54.2&11.6 \\
  SOLOv2 &  ResNet-$101$ &3x &39.7  & 60.7   & 42.9  & 17.3  &42.9  & 57.4& 15.2\\
  CondInst &  ResNet-$101$ &3x &39.1  &  60.9 & 42.0  &21.5   & 41.7  & 50.9& 11.0 \\
  Mask2Former & ResNet-$101$ &4x &44.2 &-&-&23.8 &47.7 &66.7 &7.8\\
  DiffusionInst &  ResNet-$101$ &5x & 41.0&63.9&43.9&20.7&44.4&59.9& 1.6\\ 
  DiffusionInst(4-step) &  ResNet-$101$ &5x &41.1&64.3&44.0&20.8&44.4&59.8 &1.6\\ \midrule
  %Mask RCNN &  Swin-T & 1x & 39.8 & 63.3 & 42.7 & 24.2 & 43.1  & 54.6 \\
  %Mask RCNN & Swin-S & 1x & 41.1 & 64.9 & 44.2 & 25.1 & 44.3 & 56.6 \\
  %Mask RCNN &  Swin-B & 1x & 42.3  & -  & -  & -  & -  & - \\
  %Mask RCNN &  Swin-T & 3x & 41.6  & 65.1  & 44.9  & 25.9 & 45.1 & 56.9 \\
  %Mask RCNN & Swin-S & 3x & 43.3 & 67.3 & 46.6 & {28.1} & 46.7  & 58.6& -\\
  Mask RCNN &  Swin-B & 3x & 43.4  & 66.8  & {46.9}  & -  & -  & - &-\\
  %Cascade Mask RCNN &  Swin-T & 3x & 43.7 & 66.6 & 47.3 & \textbf{28.8} & 48.7 & 60.6 &-\\ 
  QueryInst & Swin-L &4x &44.6& 68.1& 48.7& 26.6 &46.9 &57.7 &3.1\\ 
  Mask2Former & Swin-L &8x &\textbf{50.1}&-&-& \textbf{29.9}& \textbf{53.9} &\textbf{72.1} &4.0\\
  %Cascade Mask RCNN &  Swin-S & 3x & 45.0 & 68.2 & 48.8 & 28.8 & 48.7 & 60.6 \\ 
  %Cascade Mask RCNN &  Swin-B & 3x & 45.0 & 68.1 & 48.9 & 28.9 & 48.3 & 60.4 \\ 
  %Mask2Former&  Swin-L &10x & 50.1 & -& -&29.9 &53.9 &72.1 \\
  DiffusionInst &  Swin-B &5x & 46.6&{71.4}&{50.2}&26.7&50.5&67.1& 1.8 \\
  %DiffusionInst(4-step) &  Swin-B &5x &\textbf{46.8}&\textbf{71.7}&\textbf{50.3}&26.7&\textbf{50.7}&\textbf{67.1}&1.7\\
  DiffusionInst &  Swin-L &5x &47.8&\textbf{72.8}&\textbf{51.9}&28.4&51.7&67.8&  1.2 \\
  DiffusionInst(4-step) &  Swin-L &5x& \textbf{47.8}&\textbf{73.0}&\textbf{51.8}&\textbf{28.6}&\textbf{51.7}&\textbf{67.8}&1.1 \\
 \bottomrule
 \end{tabular}%}
\caption{\textbf{Results (AP\%) of instance segmentation on COCO.} We list the performance of existing popular instance segmentation approaches on different backbones. For a fair comparison, models are trained using only COCO training data. Among them, our DiffusionInst achieves competitive performances, especially with large backbones like Swin-B and Swin-L. Top 2 results are in bold. We also show the FPS measured on a single V100 GPU with batch size 1 during inference for fair.}
\label{sota}
\end{table*}

There are two advantages of using filters to represent instance masks in the diffusion process. One is directly denoise random noise to a whole mask figure is much more complicated than a vector. While DiffusionDet has shown marvelous results in the noise-to-box setting, it is natural to propose a noise-to-filter process with its success. Another benefit is that we replace the widely used box to mask prediction scheme, \emph{i.e.}, decoding RoI features to local masks, with the dynamic mask head for predicting global masks. Unlike bounding boxes, we believe instance masks need larger receptive fields due to the higher requirements on instance edges. The RoI features are usually cropped from downsampled feature maps, in which sizes the details of instance edges are all missing. To this end, representing masks as a combination of filters and the multi-scale feature can help us to build DiffusionInst with satisfactory instance segmentation performances. %We will introduce 

\subsection{DiffusionInst}
%%%%%%%%%%%%%%%%%%%%%%%%%%%%%%%%%%%
With the above mask representation method from CondInst, we can regard a data sample in DiffusionInst as a filter $\mathbf{x}_0=\bm{\theta}$ for instance segmentation. The overall framework of the DiffusionInst is illustrated in Figure~\ref{framework}. The whole architecture mainly contains the following components: (1) A CNN (\emph{e.g.} ResNet-50\cite{ResNet}), or Swin (\emph{e.g.} Swin-B\cite{swin}) backbone is utilized to extract compact visual feature representations with FPN\cite{FPN}. (2) A mask branch is utilized to fuse different scale information from FPN, which outputs a mask feature $\mathbf{{F}}_{mask}\in \mathcal{R}^{c\times H/4\times W/4}$. These two components work like an encoder, and the input image will only pass them once for feature extraction. (3) As for the decoder, we take a set of noisy bounding boxes associated with filters as input to refine boxes and filters as a denoise process. This component is borrowed from DiffusionDet and can be iteratively called. (4) Finally, we reconstruct the instance mask with the help of mask feature $\mathbf{{F}}_{mask}$ and denoised filters. Like DiffusionDet, we keep its optimization targets on bounding boxes but omit them here for better understanding. 
%%%%%%%%%%%%%%%%%%%%%%%%%%%%%%%%%

\noindent \textbf{Training:} During training, we tend to construct the diffusion process from groundtruth to noise filters relying on the corresponding bounding boxes. After adding the noise, we train the model to reverse this process. Assuming an input image has $N$ instance masks ($\mathbf{m}^{gt}\in \mathcal{R}^{N\times H \times W}$) need to be segmented. We randomly choose a time $t$ to perturb these groundtruth boxes to noisy ones with Equation~\ref{eqn.forward}. Noisy instance filters for training are also generated with noisy box features and one fully-connected layer denoted as $\eta$. The details for groundtruth padding and corruption can be found in DiffusionDet. In conclusion, we can obtain the predicted instance masks as (the denoise process of the decoder is denoted as $f(\bm{b},t)$): 
\begin{equation}
\begin{split}
\bm{b}_t &= \sqrt{\bar{\alpha_t}}\bm{b}^{gt}_0+(1-\bar{\alpha_t})\mathbf{\epsilon}, \\
\bm{\theta}_0 &= \eta(f(\bm{b}_t,t)), \\
%\bm{\theta}_{t-1} &= f(\bm{\theta}_t,t), \\
\mathbf{m} &= \mathcal{\phi}(\mathbf{{F}}_{mask};\bm{\theta}_0).
\end{split}
\end{equation}
With the dice loss\cite{milletari2016v} used in CondInst, we can obtain the training objective function as:
\begin{equation}
L_{overall}= L_{det} + \lambda L_{dice}(\mathbf{m},\mathbf{m}^{gt}),
\end{equation}
where $L_{det}$ is the training loss of DiffusionDet and $\lambda$ being $5$ in this work is used to balance the two losses. Following DiffusionDet, we perform multiple supervisions from different decoder stages on mask losses.

\noindent \textbf{Inference:} The inference pipeline of DiffusionInst is a denoising sampling process from noise to instance filters. Starting from boxes $\bm{b}_T$ sampled in Gaussian distribution, and the model progressively refines its predictions as follows:
\begin{equation}
\begin{split}
\bm{b}_{0} &= f(\cdots (f(\bm{b}_{T-s},T-s))) \quad s=\{0,\cdots,T\}, \\
\bm{\theta}_{0} &= \eta(\bm{b}_{0}), \\
\mathbf{m} &= \mathcal{\phi}(\mathbf{{F}}_{mask};\bm{\theta}_{0}).
\end{split}
\end{equation}
Note that DDIM is also used in our model following DiffusionDet.

\subsection{Discussions}\label{discuss}
Although we have successfully introduced the diffusion model into the instance segmentation task, some aspects still require improvements. The first thing is that our noise-to-filter process still relies on the bounding boxes due to the difficulty of obtaining groundtruth filters. In the future, we would like to see whether we can directly train our DiffusionInst without objective functions from bounding boxes. %We are now working on one possible solution, \emph{e.g.}, attention mechanism. 

Secondly, a more significant performance gain of multi-step denoising is needed. Specifically, when performing 4-step denoising, it only improves less than 1\% AP. In the future, we would like to explore a new sample strategy instead of DDIM, leading to more effective multi-step denoising.

Thirdly, since diffusion models are naturally proposed to tackle generative tasks, the noise-to-filter process in discriminative tasks needs more accurate instance contexts as the condition. Instance contexts rely heavily on the representative backbone features and large receptive fields, whose architectures are still open to researchers.

Finally, DiffusionInst takes more epochs to get satisfactory performances than standard instance segmentation approaches such as SOLO and Mask RCNN. During inference, the speed of DiffusionInst is also slower than theirs. How to design a faster training scheme and more efficient denoise process are essential but still unexplored.

\begin{table}[t]
\centering
\resizebox{0.48\textwidth}{!}{
 \begin{tabular}{c|c|c|ccc}
\toprule 
  \multirow{2}{*}{Methods}
  & \multirow{2}{*}{Backbone}
  & \multicolumn{4}{c}{COCO} \\
  & 
  & AP 
  & AP$_{\mathtt{S}}$ 
  & AP$_{\mathtt{M}}$ 
  & AP$_{\mathtt{L}}$  \\
  \midrule
  Mask RCNN & ResNet-$50$ & 36.8 & 17.1 & 38.7 & 52.1\\
  %YoLACT & ResNet-$50$ & 28.2 & 9.2 & 29.3 & 44.8\\
  %PolarMask & ResNet-$50$ & 29.1 & 12.6 & 31.8 & 42.3\\
  SOLOv2& ResNet-$50$ & 38.2 & 16.0 & 41.2 & 55.4 \\
  CondInst& ResNet-$50$ & 37.8 & 18.2 & 40.3 & 52.7 \\
  SOLQ & ResNet-$50$ & 39.7& 21.5 &42.5 &53.1 \\
  QueryInst & ResNet-$50$ & 39.9 & 22.9 &41.7 &51.9 \\
  DiffusionInst& ResNet-$50$ & 37.1 & 19.4 &39.7  &49.3  \\
   \midrule
  Mask RCNN &  ResNet-$101$  & 38.3 & 18.2 & 40.6 & 54.1\\
  %YoLACT & ResNet-$101$ & 31.2 & 12.1 & 33.3 & 47.1 \\
  %SOLO & ResNet-$101$ & 37.8 & 16.4 & 40.6 & 54.2 \\
  SOLOv2& ResNet-$101$ & 39.7 & 17.3 & 42.9 & 57.4 \\
  %PolarMask & ResNet-$101$ & 32.1 & 14.7 & 33.8 & 45.3\\
  %EmbedMask& ResNet-$101$ & 37.7 & 17.9 & 40.4 & 53.0\\
  CondInst& ResNet-$101$ & 39.1 & 21.5 & 41.7 & 50.9\\
  SOLQ&ResNet-$101$ &40.9& 22.5& 43.8 &54.6 \\
  QueryInst & ResNet-$101$ & 41.7& 24.2& 43.9& 53.9  \\
  DiffusionInst& ResNet-$101$ &41.5  &22.9  &44.3  & 55.2 \\
 \midrule
 %SOLQ& Swin-L & 46.7 & 29.2 & 50.1 & 60.9  \\
 SOLQ & Swin-L & 46.7& 29.2& 50.1& 60.9  \\
   QueryInst & Swin-L & \textbf{48.9} & \textbf{30.8} &\textbf{52.6}& \textbf{68.3}  \\
   %Mask2Former&Swin-L &\textbf{50.5} &29.1	&\textbf{53.8}	&\textbf{71.2} \\
  DiffusionInst & Swin-B & 47.6 & {28.3} & {50.5} & {63.4} \\
  DiffusionInst & Swin-L & \textbf{48.3}&\textbf{29.6}&\textbf{51.5}&\textbf{64.0} \\
 \bottomrule
 \end{tabular}}
\caption{\textbf{Results (AP\%) of instance segmentation on COCO test-dev dataset.} We use 100 predefined filters and perform a one-step denoise according to its online evaluation strategy. The top 2 results are in bold.}
\label{sota_cocotestdev}
\end{table}

\begin{table}[t]
\centering
 \begin{tabular}{c|c|c|ccc}
\toprule 
  \multirow{2}{*}{Methods}
  & \multirow{2}{*}{Backbone}
  & \multicolumn{4}{c}{LVIS} \\
  %& \multicolumn{6}{c}{LVIS} \\ 
  & 
  & AP 
  & AP$_{\mathtt{r}}$ 
  & AP$_{\mathtt{c}}$ 
  & AP$_{\mathtt{f}}$  \\
  %& AP 
  %& AP$_{\mathtt{50}}$ 
  %& AP$_{\mathtt{75}}$ 
  %& AP$_{\mathtt{S}}$ 
  %& AP$_{\mathtt{M}}$
  %& AP$_{\mathtt{L}}$ \\
  \midrule
  Mask RCNN &  ResNet-$50$  &16.1  &0.0 & 12.0  & 27.4 \\
  +EQL &  ResNet-$50$  &18.6   & 2.1& 17.4 &27.2      \\
  +RFS &  ResNet-$50$  &22.2 &11.5& 21.2& 28.0 \\
  +EQLv2 &  ResNet-$50$  &25.5 &17.7& 24.3 &30.2 \\
  DiffusionInst &  ResNet-$50$  & 22.3 &13.9&20.7&27.0  \\ \midrule
  Mask RCNN &  ResNet-$101$  & 21.7& 1.6& 20.7& 31.7 \\
  +RFS &  ResNet-$101$  &25.7& 17.5& 24.6& 30.6   \\
  +EQLv2 &  ResNet-$101$  &27.2 &20.6& {25.9}& 31.4  \\
  DiffusionInst &  ResNet-$101$  & 27.0 & 19.7 & 25.9 &31.5    \\ \midrule
  Mask RCNN & Swin-T& 28.6&-&-&- \\
  DiffusionInst &  Swin-B  &{36.0}  & {28.7} & {35.7} & {39.5}  \\ \bottomrule
  %DiffusionInst &  Swin-L  &  \\ \bottomrule
 \end{tabular}
\caption{\textbf{Results (AP\%) of instance segmentation on LVIS.} We report the performances of our DiffusionInst (schedule 3x, one-step) and three advanced models built on Mask RCNN. The best results are in bold.}
\label{sota_lvis}
\end{table}
\begin{table}[thb]
%\resizebox{0.49\textwidth}{!}{ 
\centering
\begin{tabular}{c|c|c}
\toprule
\multicolumn{2}{c|}{Architectures}           & AP  \\ \hline
\multirow{5}{*}{Mask Loss Weight}  & $\lambda = 1$    &   34.1   \\
& $\lambda = 5$  & 35.1 \\ 
& $\lambda = 10$  &  34.5\\ 
& multi-stage & 34.2 \\ 
& $\lambda = 5$ \& multi-stage & \textbf{37.3} \\ \hline
\multirow{4}{*}{\# Mask Feature Channel}  & $c=1$    &  33.8    \\
& $c=4$ & 36.9 \\ 
& $c=8$ & \textbf{37.3} \\ 
& $c=16$ & 37.1 \\ \hline
\multirow{4}{*}{\# Mask Head Layer}  & 1 ($d=9$) &  33.2 \\
& 2 ($d=81$)  &     36.8   \\
& 3 ($d=153$)  &     \textbf{37.3}   \\
& 4 ($d=225$)  &     \textbf{37.3}  \\ \hline
\multirow{4}{*}{\# Predefined Filter}  & 100 & 33.4  \\
& 300  &   36.7     \\
& 500  &    37.3  \\
& 1000  &    \textbf{37.4}    \\
\bottomrule
\end{tabular}
%}
\caption{\textbf{Architecture variants of DiffusionInst on COCO.} We evaluate different architecture variants of DiffusionInst from several views, including different weights for mask loss, the number of channels, layers and filters. Note that we use ResNet-$50$ as the backbone (schedule 5x, one-step denoise).}
\label{tab:ablation}
\end{table}

\section{Experiments}

\subsection{Datasets}

We conducted extensive experiments on two standard
instance segmentation datasets: COCO\cite{COCO} and LVIS\cite{LVIS}. For all two datasets, we used the standard mask AP metric\cite{COCO} as the evaluation metric.

\noindent
\textbf{COCO.} COCO is an 80-category label set with instance-level annotations. Following\cite{PointRend}, we use the COCO train2017 (118K training images) for training, and the ablation study is carried out on the val2017 (5K validation images). We also report our main results on test-dev (20k images) for comparison.

\noindent
\textbf{LVISv1.0.} We further perform experiments on a more challenging LVIS dataset\cite{LVIS}. LVIS is a long-tail instance segmentation dataset containing 1203 categories, having more than 2 million high-quality instance mask annotations. LVIS contains 100k, 19.8k, and 19.8k images for training, validation, and testing. According to the frequency of occurrence in the training set, the categories are divided into three groups: rare (1-10 images), common (11-100 images), and frequent ($>$100 images). We also report them as AP$_{\mathtt{r}}$, AP$_{\mathtt{c}}$, AP$_{\mathtt{f}}$.

\subsection{Implement Details}

In our experiments, we choose the ResNet-$50$\cite{ResNet}, ResNet-$101$, Swin-Base\cite{swin} and Swin-Large with FPN\cite{FPN} as the backbone in the proposed method. Note that the Swin transformer backbones are pretrained on ImageNet22k with resolution $224\times 224$. We implement the proposed method with PyTorch\cite{pytorch} and it takes about 26 hours to train a DiffusionInst (ResNet-$50$) on 8 A100 GPUs with batch size 32. The optimizer of the proposed method is AdamW\cite{loshchilov2017decoupled}, with a learning rate of 2.5e-5 and a weight decay of 1e-4. Following DiffusionDet, standard data augmentation strategies contain random horizontal flip, scale jitter, and random crop augmentations. Other substantial data augmentation like MixUp\cite{zhang2017mixup} or Mosaic\cite{ge2021yolox} are not used.

\subsection{Comparison with State-of-the-art}

In this section, we experiment with our DiffusionInst in three datasets: COCO validation, COCO test-dev and LVISv1.0. We compare our model with popular existing methods such as Mask RCNN, Cascade Mask RCNN, SOLO, CondInst, Mask2Former, SOLQ and QueryInst on various backbones. Note that the performances in this section are all obtained without pretraining on extra detection data like Objects-365\cite{2020Objects365}.

\noindent\textbf{COCO validation set.} In Table~\ref{sota}, we list seven existing instance segmentation models' performances with various backbones and schedules. To better understand the results, we highlight the top 2 results. When using Swin-Large as the backbone, DiffusionInst achieves top 2 performance beyond a strong baseline, \emph{i.e.}, QueryInst. Mask2Former has performed more training iterations for the best results in this table. We can draw the following conclusions from this table: 

Firstly, as the backbone complexity and capacity increase, the performance gains are enlarged. For example, our DiffusionInst increases about 9\% AP when changing ResNet-$50$ to Swin-Base. This observation means the diffusion model needs more representative features outputted from a stronger backbone as the condition for the diffusion process since we do not have inductive bias. In other words, instance-aware features are essential for denoise as the condition for box and filter refinement. 

Secondly, removing the RPN also leads to slower convergence (usually 3x or 5x schedule) since the model has to find instance locations by itself. For example, taking ResNet-$50$ as the backbone, DiffusionInst with 1x training iterations only achieves 30.4\% AP, which is 4\% AP smaller than Mask RCNN. When we finish 3x training, the difference is narrowed to 2\% AP.

Thirdly, multi-step denoise has incremental benefits (usually less than 0.3\% AP), but as a diffusion model, it naturally has a smaller FPS than existing methods. We also have to claim that the FPS numbers in Table~\ref{sota} are evaluated on a single GPU V100. On single A100, DiffusionDet achieves 30 FPS as reported in \cite{DiffusionDet}, while our DiffusionInst also has 15 FPS. 

Finally, DiffusionInst performs better on large instances (AP$_{L}$) but sometimes misses small ones, which indicates that the diffusion model needs larger receptive fields on features in the instance segmentation.

\noindent\textbf{COCO test-dev set.} 
Note that COCO test-dev only evaluates the top 100 predicted instances. Thus, we only employ 100 predefined filters in DiffusionInst in Table~\ref{sota_cocotestdev}. Similar conclusions can be drawn from the COCO validation set. That is, DiffusionInst shows steady improvement when the backbone size scales up. When equipped with ResNet-$50$, it achieves 37.1\% AP, which is smaller than all five baselines in Table~\ref{sota_cocotestdev}. When using ResNet-$101$, DiffusionInst achieves almost the best performance in the same setting except for the 0.2\% AP gap with QueryInst. Finally, when DiffusionInst utilizes ImageNet-22k pre-trained Swin-Large as the backbone, it obtains 48.3\% AP, outperforming a strong baseline method SOLQ.

\begin{figure*}[t!]
\centering
\includegraphics[width=\textwidth]{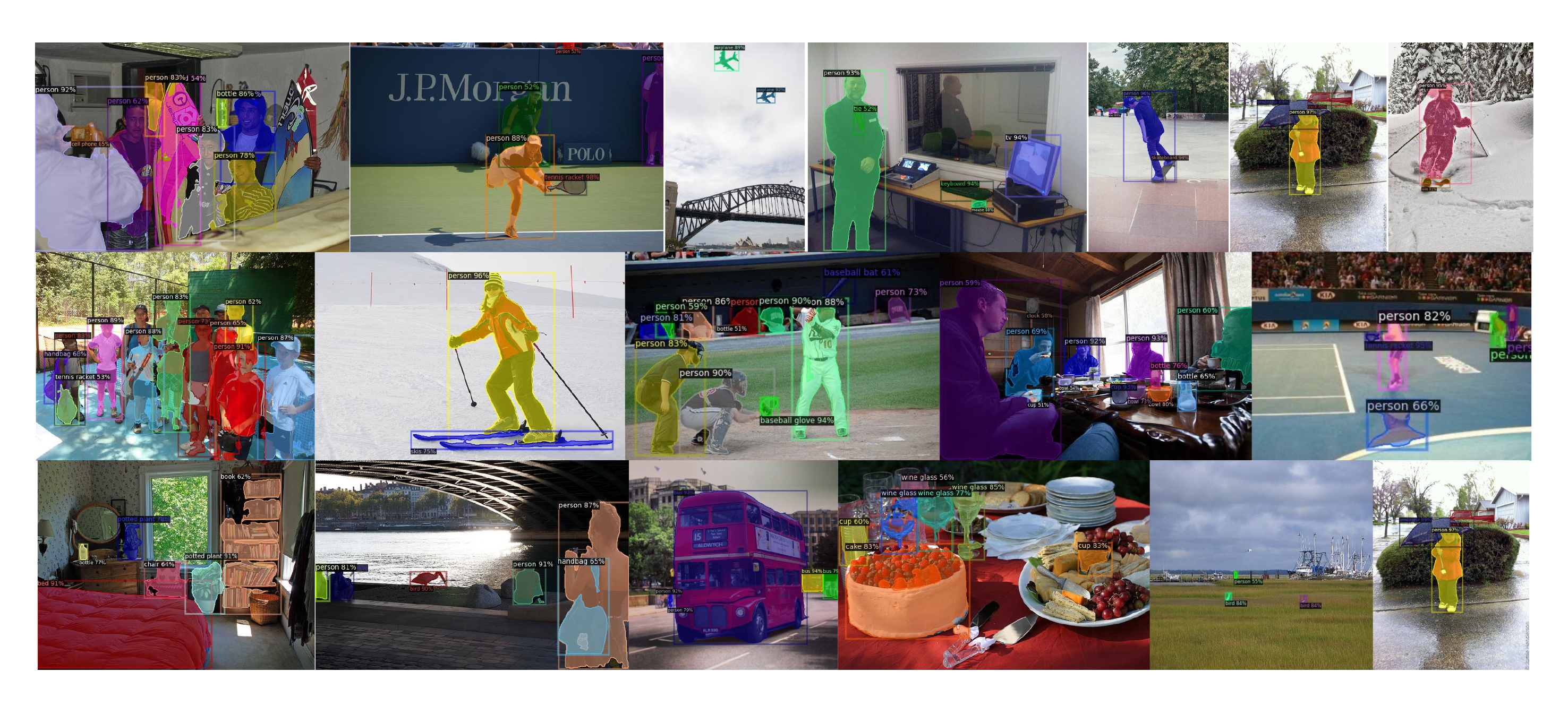}
\caption{\textbf{Visualization of our DiffusionInst on COCO validation set.} Note that the model is based on ResNet-$50$ and one-step denoising.}
\label{visualizations}
\end{figure*}

\noindent\textbf{LVIS dataset.} This dataset uses the same images with COCO but pays more attention to long-tail instances. Existing approaches on this dataset main built on Mask RCNN, such as EQL\cite{tan2020equalization}, RFS\cite{LVIS} and EQLv2\cite{tan2021equalization}. Our DiffusionInst achieves the best performance with Swin-Base as the backbone. Moreover, our DiffusionInst attains more remarkable gains in this dataset on Mask RCNN. For example, DiffusionInst surpasses Mask RCNN 0.3\% AP with ResNet-$50$ as the backbone on COCO test-dev dataset but enlarge its advantage on LVIS, demonstrating that our noise-to-filter denoise process would become more helpful for a more challenging benchmark.

\subsection{Ablation Studies}

We conduct experiments on the COCO validation set with the ResNet-$50$ backbone, 5x schedule and one-step denoising for the ablation studies as shown in Table~\ref{tab:ablation}. In this table, we evaluate different architecture variants of DiffusionInst from several views, including different weights for mask loss, the number of channels, layers and filters.

\noindent\textbf{Mask Loss Weights.} Since DiffusionInst is trained via more than one loss function. It is natural to balance them with different loss weights. As shown in the first five rows of Table~\ref{tab:ablation}, we vary mask loss weight $\lambda$ from $1$ to $10$ and find $\lambda=5$ achieving the best performance. Another observation is that employing multi-stage mask loss supervision following DiffusionDet with $\lambda=5$ can bring 2.2\% AP improvements.

\noindent\textbf{Mask Branch.} To enhance the expressiveness of the mask feature, we further explore the channel number of the mask branch. Unlike CondInst, we ignore its relative coordinate map and still get good performance. 
Among different choices, the 8-channel mask feature achieves 37.3\% AP, and extra channels cannot improve performance. We set the channel number of the mask feature to 8 by default. 

\noindent\textbf{Mask Head.} The dynamic mask head plays a critical role in our method. Thus, we conduct ablation studies to show the impact of parameters in the mask head. As presented in the table, with the number of convolutions in the mask head increasing, the performance improves steadily and achieves the peak of 37.3\% AP with three stacked convolutions. More convolutions will not contribute to the final results.

\noindent\textbf{Predefined filter numbers.} The number of predefined filters has a similar function to the number of proposals in standard instance segmentation approaches. In this table, we vary filter numbers from 100 to 1000 and choose to use 500 as a good balance of performance and model complexity.

\subsection{Visualizations}

The visualization of the proposed method on the COCO validation dataset is shown in Figure~\ref{visualizations}. In conclusion, our DiffusionInst can successfully segment instances with an accurate boundary but still miss some instances in bad cases. For example, some persons are ignored when they only occur in a small region of the whole image. Moreover, as shown in the first column, occluded instances, \emph{e.g.}, persons and books, are also easy to misclassify.

\section{Conclusion}
This work introduced a novel diffusion framework for the instance segmentation task. Regarding instance segmentation as a noise-to-filter process, our DiffusionInst achieves comparable single-model results ({\em i.e.}, Swin-Large backbone) on the COCO and LVIS datasets. However, as mentioned in Section~\ref{discuss}, we also conclude several limits of DiffusionInst, namely relies on the bounding boxes resulting in slightly poor performance on small instances, the unsatisfactory performance gain from multi-step denoising and the longer training time with slower inference speed.
Even though our model has the above weaknesses, the diffusion model is a new possible solution to this task in the future. We hope our work could serve as a strong baseline, which could inspire designing more efficient frameworks and rethinking the learning targets for the challenging instance segmentation task.

\bibliographystyle{named}
\bibliography{ijcai22}

\end{document}